\documentclass[authoryear,preprint,review,12pt]{elsarticle}

\usepackage{amssymb}

\usepackage{amsthm}
\usepackage{amsmath}
\usepackage{caption}
\usepackage{subcaption}
\usepackage{lscape}
\usepackage[table]{xcolor}
\usepackage{hyperref}
\hypersetup{
    colorlinks=true,
    linkcolor=blue,
    filecolor=magenta,      
    urlcolor=cyan,
}

\theoremstyle{definition}
\newtheorem{defn}{Definition}[section]

\newtheorem{exmp}{Example}[section]

\theoremstyle{remark}

\journal{Elsevier journal}

\begin{document}

\begin{frontmatter}



\title{A thermodynamical approach towards\\ multi-criteria decision making (MCDM)}

\author[l1,l2]{Mohit Verma\corref{c1}}
\ead{mohitverma@serc.res.in}
\cortext[c1]{Corresponding author}
\author[l1,l2]{J. Rajasankar}
\address[l1]{CSIR-Structural Engineering Research Centre, Chennai 600 113, India}
\address[l2]{Academy of Scientific and Innovative Research (AcSIR), Chennai 600 113, India}
\begin{abstract}
In multi-criteria decision making (MCDM) problems, ratings are assigned to the alternatives on different criteria by the expert group. In this paper, we propose a thermodynamically consistent model for MCDM using the analogies for thermodynamical indicators - energy, exergy and entropy. The most commonly used method for analysing MCDM problem is Technique  for  Order  of  Preference  by  Similarity  to  Ideal  Solution (TOPSIS). The conventional TOPSIS method uses a measure similar to that of energy for the ranking of alternatives. We demonstrate that the ranking of the alternatives is more meaningful if we use exergy in place of energy. The use of exergy is superior due to the inclusion of a factor accounting for the quality of the ratings by the expert group. The unevenness in the ratings by the experts is measured by entropy. The procedure for the calculation of the thermodynamical indicators is explained in both crisp and fuzzy environment. Finally, two case studies are carried out to demonstrate effectiveness of the proposed model.
\end{abstract}

\end{frontmatter}

\section{Introduction}
Multi-criteria decision making (MCDM) is a process used for ranking of alternatives based on different criteria. The applications of MCDM are numerous and it has been applied to human resource management \citep{shih2007extension}, transportation \citep{tsaur2002evaluation}, portfolio optimization \citep{ehrgott2004mcdm}, product design \citep{liu2011product}, vendor selection \citep{shyur2006hybrid} and visual inspection \citep{verma2015fuzzy}. The most commonly used method for MCDM is Technique  for  Order  of  Preference  by  Similarity  to  Ideal  Solution (TOPSIS). The advantages of TOPSIS includes \citep{shih2007extension} - scalar value accounting for both best and worst alternative; logical representation of human rationale and easy implementation. TOPSIS is based upon the concept that the chosen solution should be closest to positive ideal solution and farthest from negative ideal solution. 

The motivation for the present study comes from the application of thermodynamics in the field of bibliometric research by \cite{prathap2011energy}. The analogies of the energy, exergy and entropy energy, exergy and entropy associated with a bibliometric sequence were used to derive an indicator of a scientist’s performance. In this paper, we present a model for MCDM in the paradigm of thermodynamics. We define analogies for thermodynamical indicators - energy, exergy and entropy with respect to MCDM. It should be noted that the entropy defined in the present study is different from Shannon's entropy \citep{shannon2015mathematical} which assumes a prior distribution. A natural definition of entropy derived from the first principles is used in the present study. It is observed that the conventional TOPSIS method uses a measure similar to what we define as energy indicator. We demonstrate with the help of examples that it is exergy indicator which makes more sense in the ranking of alternative than energy indicator. The proposed model is quite simple to implement and is thermodynamically consistent. The proposed model is formulated for both crisp and fuzzy environment. The effectiveness of the proposed model is demonstrated with the help of two case studies (covering both crisp and fuzzy environment).   

The organization of the paper is as follows. The second section defines the preliminaries towards thermodynamics. In the third section, we define analogies for the energy, exergy and entropy in both crisp and fuzzy environment. The fourth section describes why using exergy indicator makes more sense than using an indicator based on energy. Fifth section lists out the procedure for MCDM using thermodynamical indicators. Afterwards, two case studies are carried out to demonstrate the effectiveness of the method in both crisp and fuzzy environment. The final section presents the conclusions drawn from the present study.

\section{Preliminaries towards thermodynamics}
Thermodynamics is viewed as the science of energy. In this section, we reproduce the definition of the terms like energy, exergy and entropy based on \cite{dincer2001energy} for the sake of completeness. The section also describes the two basic laws which govern the science of thermodynamics.
\begin{defn}
Energy ($U$) of a system is defined as its ability to do work. It can neither be created nor be destroyed but can only be converted from one form to another. It depends on the parameters of the matter or energy flow only and is independent of environment parameters. It is a measure of quantity alone.
\end{defn}

\begin{defn}
Exergy ($X$) of a system is the maximum useful work possible during a process that brings the system into equilibrium with the specified reference environment. Exergy is the potential of a system to cause a change as it achieves equilibrium with its environment. It depends upon parameters of matter or energy flow and environment. It is a measure of both quantity and quality.
\end{defn}

\begin{defn}
Entropy ($S$) of a system is the measurement of the amount of disorder in the system. A system can generate entropy. The entropy of the system can be increased or decreased by energy transport across the system boundary. The direction of the change in the states of the system is from a state of low probability to the one with higher probability. Since, the disordered states are more probable than ordered states, the natural direction of the change in system states is from order to disorder. 
\end{defn}
\noindent
\textbf{First law of thermodynamics}\\ The energy is a thermodynamic property which can change from one form to another but the total amount of energy remains constant. It is based on the conservation of energy. \\[3mm]
\noindent
\textbf{Second law of thermodynamics}\\ The energy has quality as well as quantity, and actual processes occur in the direction of decreasing quality of energy. Any process either increases the entropy of the universe - or leaves it unchanged. \\[3mm]
\noindent
The first law of thermodynamics gives no information about the direction of the energy conversion or the quality of energy. It is the second law of thermodynamics which establishes the difference in the quality of the various forms of energy. Based on second law of thermodynamics, entropy can be seen as the measure of energy which is unavailable for direct conversion to work. Two systems can have same energy but may not able capable of doing the same amount of useful work. A system which is capable doing more useful work is said to have good quality of energy compared to other.

\section{Thermodynamical analogies}
In this section, we define analogies for the thermodynamical terms in both crisp and fuzzy environment. These analogies lay down the basis for the thermodynamically consistent MCDM model. Let an alternative ($A$) is rated by a decision maker ($E$), for a criterion ($C$). The weight assigned to the criterion by the expert is $w$. The rating and the weights are normalized between 0 to 1. The rating and the weights are expressed as fixed numbers ($r,w$) in case of crisp and triangular fuzzy number ($\tilde{r},\tilde{w}$) in case of fuzzy environment. A triangular fuzzy number ($\tilde{x}$) is determined by a triplet ($a,b,c$) (Fig. \ref{tfn}) whose membership function is given by:
\begin{figure}
\centering
\includegraphics[trim = 0cm 18cm 8cm 5cm, clip, width=1\textwidth]{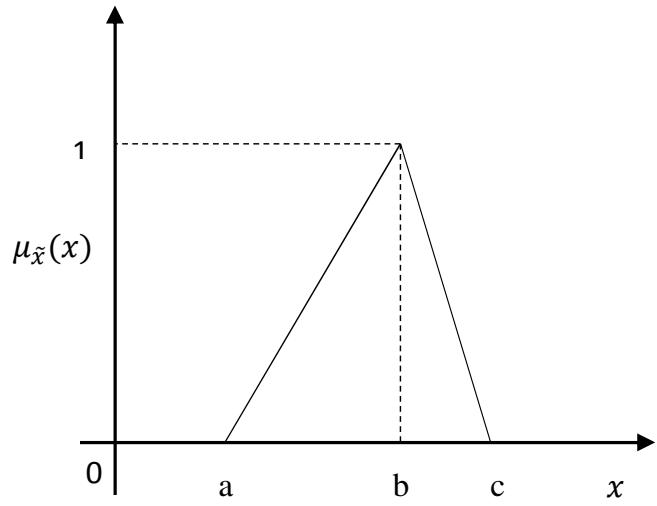}
\caption{Triangular fuzzy number}
\label{tfn}
\end{figure}
\begin{equation}
\mu_{\tilde{x}}(x)=\left \{ 
\begin{array}{rcl}
\displaystyle \frac{x-a}{b-a}&&a \leq x \leq b\\
\displaystyle \frac{x-c}{b-c}&&b \leq x \leq c\\
0&&\text{Otherwise} 
\end{array} \right.
\end{equation}
We assume that the fuzzy number associated with the rating $\tilde{r}$ is ($r_a,r_b,r_c$) and with the weight $\tilde{w}$ is ($w_a,w_b,w_c$).
\begin{figure}
\centering
\includegraphics[trim = 0cm 22.87cm 16.17cm 2.2cm, clip,width=0.3\textwidth]{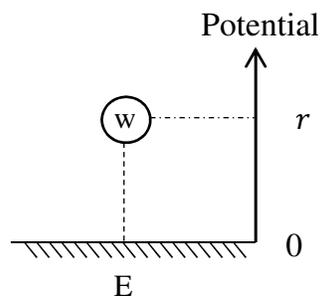}
\caption{Energy equivalence in MCDM}
\label{energy_MCDM}
\end{figure}

\begin{defn}
The \textbf{force} due to gravity or weight of an alternative in MCDM is defined as the weight assigned to it by an expert. In Fig. \ref{energy_MCDM}, the force/weight associated with alternative A is $w$.  
\end{defn}
\begin{defn}
The \textbf{potential} of an alternative in MCDM is defined as the rating assigned to it by an expert. In Fig. \ref{energy_MCDM}, the potential associated with alternative A is $r$.  
\end{defn}
\begin{defn}
The \textbf{potential difference} between two states $r_1$ and $r_2$ of the system is given by:
\begin{subequations}
\begin{align}
d&=|r_1-r_2|\\
\tilde{d}&=(|r_{1a}-r_{2a}|,|r_{1b}-r_{2b}|,|r_{1c}-r_{2c}|)
\end{align}
\end{subequations}
\end{defn}
\begin{defn}
\textbf{Work} ($W$) done by the system during the change in state from $r_1$ to $r_2$ is equal to the change in its potential energy. The work done is given by:
\begin{subequations}
\begin{align}
W&=w.d(r_1,r_2)\\
\tilde{W}&=\tilde{w}.\tilde{d}(\tilde{r_1},\tilde{r_2})
\end{align}
\end{subequations} 
\end{defn}
\begin{defn}
\textbf{Energy indicator} ($U$) of an alternative is defined as the energy possessed by virtue of its rating in the system. The energy associated with alternative A in crisp and fuzzy environment is given by:
\begin{subequations}
\begin{align}
U&=w.r\\
\tilde{U}&=\tilde{w}.\tilde{r}\\
&=(w_a,w_b,w_c).(r_a,r_b,r_c) \nonumber\\
&=(w_ar_a,w_br_b,w_cr_c) \nonumber
\end{align}
\end{subequations}
\end{defn}
\begin{defn}
The \textbf{quality} of a rating is the measure of its degree of excellence compared to other rating. If all the experts have a consensus in the rating, then quality is equal to one. It is measured as one minus the relative distance of a rating from the mean rating. Let an alternative A be rated ($r_1,r_2,...,r_n$) by $n$ experts and the mean rating is $\bar{r}$. The quality of $i^{th}$ rating is given by:
\begin{subequations}
\begin{align}
q&=\Big(1-\frac{d(r_i,\bar{r})}{\bar{r}}\Big)\\
\tilde{q}&=\Big((1,1,1)-\frac{\tilde{d}(\tilde{r_i},\bar{\tilde{r}})}{\bar{\tilde{r}}}\Big)\\
&=\Big(1-\frac{|r_{ia}-\bar{r}_{a}|}{\bar{r}_{a}},1-\frac{|r_{ib}-\bar{r}_{b}|}{\bar{r}_{b}},1-\frac{|r_{ic}-\bar{r}_{c}|}{\bar{r}_{c}}\Big)\nonumber
\end{align}
\end{subequations}
\end{defn}
\begin{defn}
\textbf{Exergy indicator} of a rating is the measure of the quality energy that a rating carries. Mathematically, it is given by:
\begin{subequations}
\begin{align}
X&=q.U\\
\tilde{X}&=\tilde{q}.\tilde{U}
\end{align}
\end{subequations}
\end{defn}
\begin{defn}
\textbf{Entropy indicator} is a measure of the unevenness in the ratings of an alternative. If an alternative is assigned exactly same rating by all the experts, then the entropy is equal to zero. Thus, the entropy of a rating can be defined as \citep{prathap2011energy}:
\begin{subequations}
\begin{align}
S&=U-X\\
\tilde{S}&=\tilde{U}-\tilde{X}
\end{align}
\end{subequations}
\end{defn}
\section{Energy vs. Exergy}
Let us assume that there are $K$ decision makers, rating $m$ alternatives based on $n$ criteria. In the classical TOPSIS method, the ratings and the weights are first aggregated using arithmetic mean or any other suitable method. The aggregated ratings and weights are then assembled to form decision ($D$) and weight ($W$) matrix as given below:
\begin{equation}
\begin{split}
D &= \left [
\begin{array}{cccc}
x_{11}&x_{12}&\cdots&x_{1n}\\
x_{21}&x_{22}&\cdots&x_{2n}\\
\vdots&\vdots&\cdots&\vdots\\
x_{m1}&x_{m2}&\cdots&x_{mn}
\end{array} \right ]\\
W &=  \left [ \begin{array}{cccc}w_1& w_2&\cdots&w_n\end{array} \right ]
\end{split}
\end{equation}
where $x_{ij}$ denotes the aggregate rating of $i^{th}$ alternative for $j^{th}$ criterion and $w_j$ represents the weight for $j^{th}$ criterion. 

In order to bring various criteria on a comparable scale, vector or linear normalization is carried out. Normalized decision matrix ($R$) given by:
\begin{equation}
R=
\left [
\begin{array}{cccc}
r_{11}&r_{12}&\cdots&r_{1n}\\
r_{21}&r_{22}&\cdots&r_{2n}\\
\vdots&\vdots&\cdots&\vdots\\
r_{m1}&r_{m2}&\cdots&r_{mn}
\end{array} \right ]
\end{equation}
where $r_{ij}$ denotes the normalized rating of $i^{th}$ alternative for $j^{th}$ criterion. The weighted normalized decision matrix is constructed by multiplying the weights with the normalized rating and is given by:
\begin{equation}
\begin{split}
V &= \left [ \begin{array}{c}
 v_{ij}\end{array} \right ],\\
 \text{where } &v_{ij}=w_j(.)r_{ij}\\
 &i = 1,2,\cdots,m\\
 &j=1,2,\cdots,n
\end{split}
\end{equation}
On careful observation, it is found that the term $v_{ij}$ is similar to what we have defined as energy in the previous section. The energy indicator associated with a rating gives the information only about the quantity and not the quality. In the process of aggregation of rating, the information on its quality is lost. There is a need for an indicator which not only accounts for the quantity but quality as well. Exergy indicator defined in the previous section includes both the quantity and the quality of the energy. This motivates the use of exergy indicator in place of energy. In this section, we highlight how the use of exergy indicator instead of energy makes more sense using two different examples covering both crisp and fuzzy environment.  
\begin{exmp}
\label{exmp1}
Consider a case where two alternatives ($A_1$ and $A_2$) are rated by 10 decision makers on a particular criterion. The aggregated weight ($w$) for the criterion is $0.7$. The normalized ratings($r$) for $A_1$ are clustered and varies from $0.4-0.6$ with a mean of $0.5$. In case of $A_ 2$, the normalized ratings are more dispersed (varies from $0.1-0.8$) but has the same mean as $A_1$ (that is $0.5$). The histogram of the normalized ratings is plotted in Fig.~\ref{ex1sc1} for $A_1$ and in Fig.~\ref{ex1sc2} for $A_2$. The normalized ratings assigned to an alternative and the calculated thermodynamical indicators are given in Table~\ref{tabex1sc1} for $A_1$ and Table~\ref{tabex1sc2} for $A_2$.
\begin{figure}
        \centering
        \includegraphics[width=0.7\textwidth]{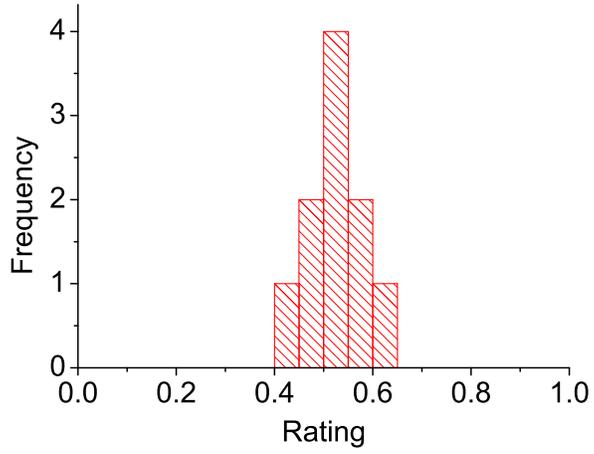}
        
    \caption{Histogram of normalized rating assigned to $A_1$ for Example \ref{exmp1}}
    \label{ex1sc1}
\end{figure}
\begin{figure} 
        \centering
        \includegraphics[width=0.7\textwidth]{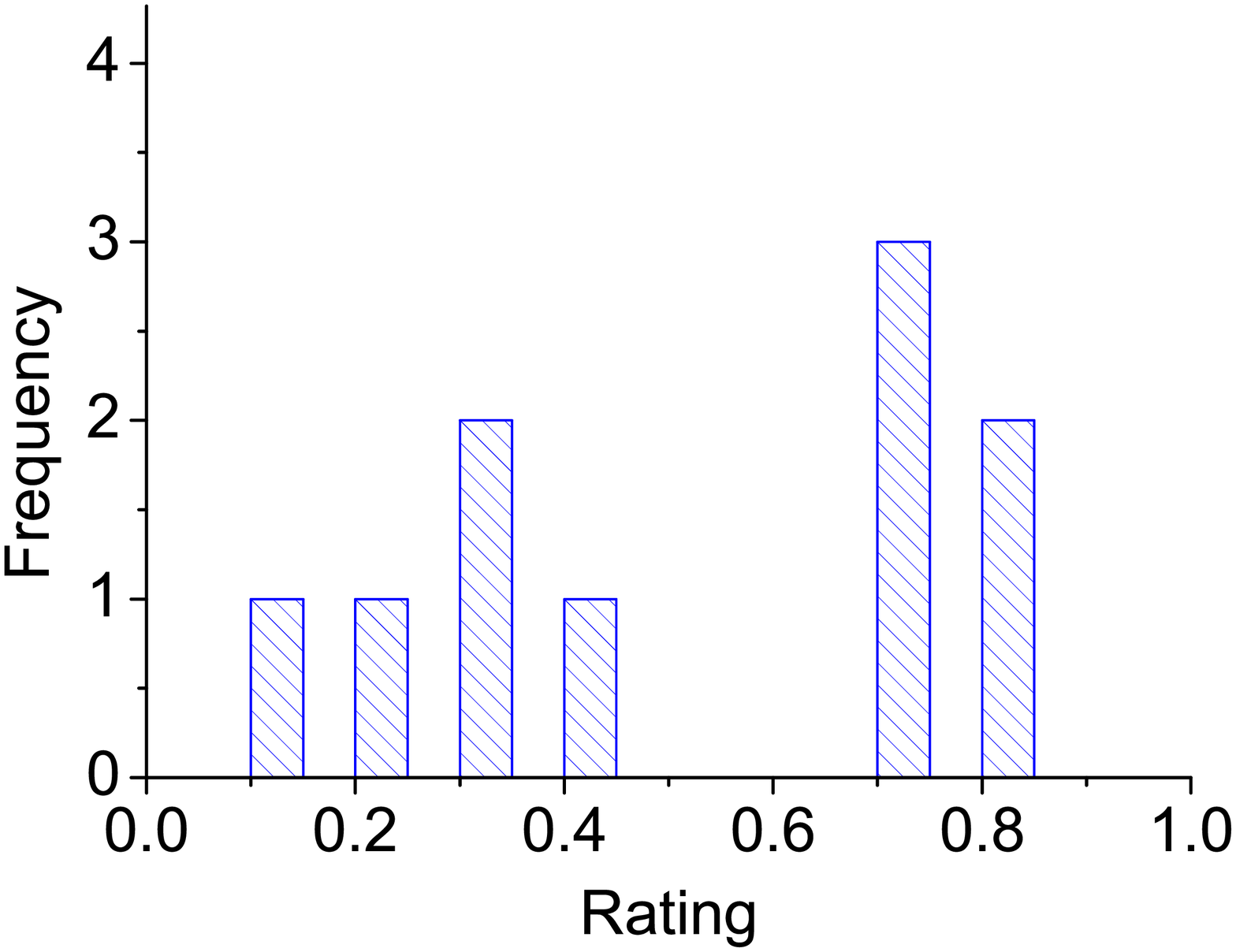}
        
    \caption{Histogram of normalized rating assigned to $A_2$ for Example \ref{exmp1}}
    \label{ex1sc2}
\end{figure}

\begin{table}
\caption{Normalized ratings and thermodynamical indicators of $A_1$ for Example \ref{exmp1}}
\label{tabex1sc1}
\centering
\begin{tabular}{cccccc}
\hline
\textbf{Decision maker}&\textbf{Rating}&\textbf{Energy}&\textbf{Quality}&\textbf{Exergy}&\textbf{Entropy}\\
&\textbf{$r$}&\textbf{$U=w.r$}&\textbf{$q$}&\textbf{$X=q.U$}&\textbf{$S=U-X$}\\
\hline
1&	0.500&	0.350&	1.000&	0.350	&0.000\\
2&	0.450&	0.315&	0.900&	0.284	&0.032\\
3&	0.400&	0.280&	0.800&	0.224	&0.056\\
4&	0.600&	0.420&	0.800&	0.336	&0.084\\
5&	0.450&	0.315&	0.900&	0.284	&0.032\\
6&	0.500&	0.350&	1.000&	0.350	&0.000\\
7&	0.500&	0.350&	1.000&	0.350	&0.000\\
8&	0.500&	0.350&	1.000&	0.350	&0.000\\
9&	0.550&	0.385&	0.900&	0.347	&0.039\\
10&	0.550&	0.385&	0.900&	0.347	&0.039\\ \hline
\textbf{Mean}&	\textbf{0.500}&	\textbf{0.350}&	\textbf{0.920}&	\textbf{0.322}&	\textbf{0.028}\\ \hline
\end{tabular}
\end{table}
\begin{table}
\caption{Normalized ratings and thermodynamical indicators of $A_2$ for Example \ref{exmp1}}
\label{tabex1sc2}
\centering
\begin{tabular}{cccccc}
\hline
\textbf{Decision maker}&\textbf{Rating}&\textbf{Energy}&\textbf{Quality}&\textbf{Exergy}&\textbf{Entropy}\\
&\textbf{$r$}&\textbf{$U=w.r$}&\textbf{$q$}&\textbf{$X=q.U$}&\textbf{$S=U-X$}\\
\hline
1&	0.200&	0.140&	0.400&	0.056&	0.084\\
2&	0.700&	0.490&	0.600&	0.294&	0.196\\
3&	0.300&	0.210&	0.600&	0.126&	0.084\\
4&	0.800&	0.560&	0.400&	0.224&	0.336\\
5&	0.100&	0.070&	0.200&	0.014&	0.056\\
6&	0.400&	0.280&	0.800&	0.224&	0.056\\
7&	0.700&	0.490&	0.600&	0.294&	0.196\\
8&	0.800&	0.560&	0.400&	0.224&	0.336\\
9&	0.300&	0.210&	0.600&	0.126&	0.084\\
10&	0.700&	0.490&	0.600&	0.294&	0.196\\ \hline
\textbf{Mean}&	\textbf{0.500}&	\textbf{0.350}&	\textbf{0.520}&	\textbf{0.188}&	\textbf{0.162}\\ \hline
\end{tabular}
\end{table}

 It is observed from Tables~\ref{tabex1sc1} and \ref{tabex1sc2} that the weighted normalized decision (which is equivalent to energy indicator) for both the alternatives will lead to same value of $0.35$, if we use classical TOPSIS method. Figures.~\ref{ex1sc1} and \ref{ex1sc2} clearly suggests that the ratings for $A_1$ are more reliable than for $A_2$. This fact is also evident from the value of mean quality indicator for $A_1$ and $A_2$. This information is lost if we use energy indicator. On the other hand, the exergy indicator clearly suggests that the $A_1$ is better rated than $A_2$. Mean entropy values of $A_1$ and $A_2$ indicates that the unevenness in the ratings assigned to an alternative by the decision makers is more in case of $A_2$ compared to $A_1$. 
\end{exmp}
\begin{exmp}
\label{exmp2}
In this example, the ratings ($\tilde{r}$) are assigned to the alternatives $A_1$ and $A_2$ in the form of triangular fuzzy number by 5 decision makers on a particular criterion. The weight ($\tilde{w}$) assigned to the criteria is a triangular fuzzy number $(0.7,0.8,0.9)$.
The normalized fuzzy rating assigned to $A_1$ and $A_2$ are shown in Figs.~\ref{fuz1} and \ref{fuz2}. The numbers in the bracket in Figs.~\ref{fuz1} and \ref{fuz2} represents the decision maker corresponding to that rating. The normalized fuzzy rating assigned to an alternative and the calculated thermodynamical fuzzy indicators are given in Table~\ref{tabfuz1} for $A_1$ and in Table~\ref{tabfuz2} for $A_2$. The mean normalized fuzzy rating and the mean fuzzy energy is same for $A_1$ and $A_2$. In this case also, the classical TOPSIS method will result in same weighted normalized fuzzy decision even though there is a large difference in the quality of ratings of $A_1$ and $A_2$. This difference is reflected in the mean fuzzy exergy indicator.
\begin{figure}
\centering
\includegraphics[trim = 0cm 18cm 8cm 5cm, clip, width=1\textwidth]{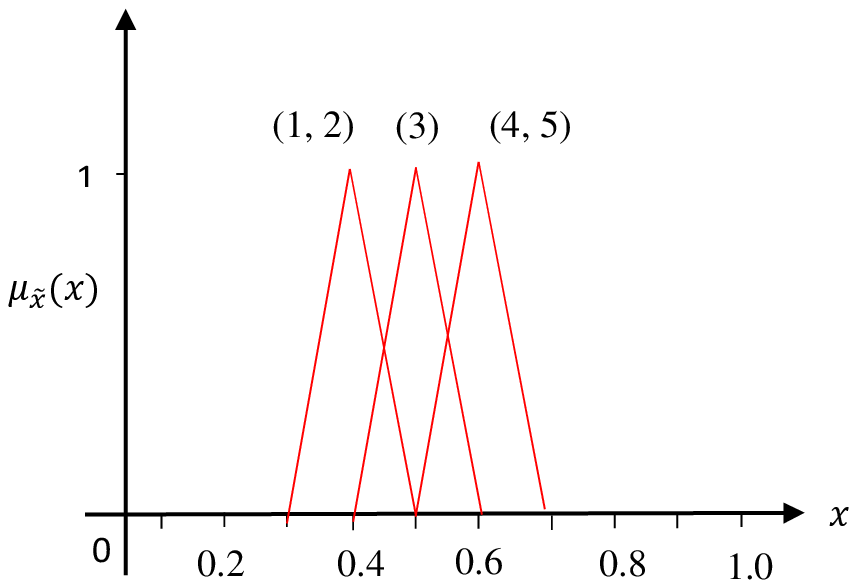}
\caption{Normalized fuzzy ratings assigned to $A_1$ for Example \ref{exmp2}}
\label{fuz1}
\end{figure}
\begin{figure}
\centering
\includegraphics[trim = 0cm 18cm 8cm 5cm, clip, width=1\textwidth]{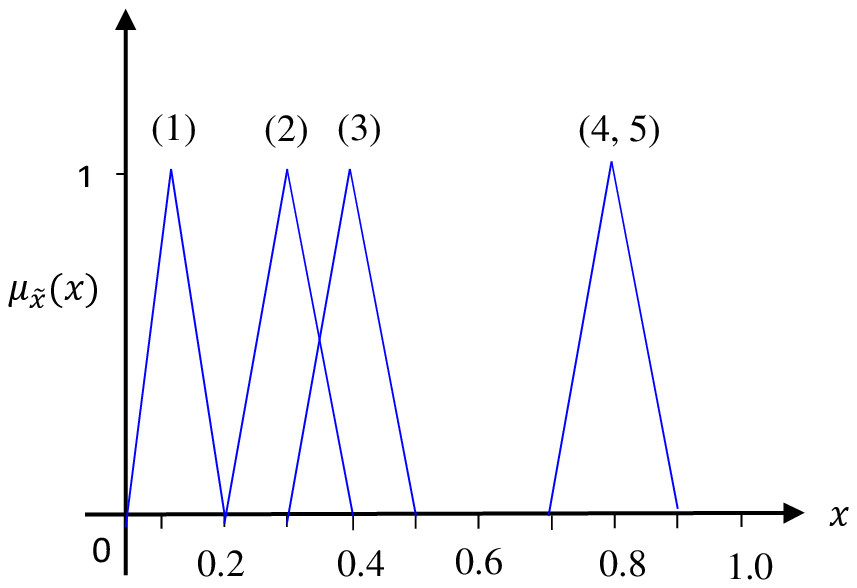}
\caption{Normalized fuzzy ratings assigned to $A_2$ for Example \ref{exmp2}}
\label{fuz2}
\end{figure}
\begin{landscape}
\begin{table}
\caption{Normalized fuzzy ratings and thermodynamical fuzzy indicators of $A_1$ for Example \ref{exmp2}}
\label{tabfuz1}
\centering
\begin{tabular}{cccccc}
\hline
\textbf{Decision maker}&\textbf{Rating}&\textbf{Energy}&\textbf{Quality}&\textbf{Exergy}&\textbf{Entropy}\\
&\textbf{$\tilde{r}$}&\textbf{$\tilde{U}=\tilde{w}.\tilde{r}$}&\textbf{$\tilde{q}$}&\textbf{$\tilde{X}=\tilde{q}.\tilde{U}$}&\textbf{$\tilde{S}=\tilde{U}-\tilde{X}$}\\
\hline
1&(0.30,0.40,0.50)&(0.21,0.32,0.45)&(0.75,0.80,0.83)&(0.16,0.26,0.38)&(0.05,0.06,0.08)\\
2&(0.30,0.40,0.50)&(0.21,0.32,0.45)&(0.75,0.80,0.83)&(0.16,0.26,0.38)&(0.05,0.06,0.08)\\
3&(0.40,0.50,0.60)&(0.28,0.40,0.54)&(1.00,1.00,1.00)&(0.28,0.40,0.54)&(0.00,0.00,0.00)\\
4&(0.50,0.60,0.70)&(0.35,0.48,0.63)&(0.75,0.80,0.83)&(0.26,0.38,0.53)&(0.09,0.10,0.11)\\
5&(0.50,0.60,0.70)&(0.35,0.48,0.63)&(0.75,0.80,0.83)&(0.26,0.38,0.53)&(0.09,0.10,0.11)\\\hline
\textbf{Mean}&\textbf{(0.40,0.50,0.60)}&\textbf{(0.28,0.40,0.54)}&\textbf{(0.80,0.84,0.87)}&\textbf{(0.22,0.34,0.47)}&\textbf{(0.06,0.06,0.07)}\\ \hline
\end{tabular}
\end{table}
\begin{table}
\caption{Normalized fuzzy ratings and thermodynamical fuzzy indicators of $A_2$ for Example \ref{exmp2}}
\label{tabfuz2}
\centering
\begin{tabular}{cccccc}
\hline
\textbf{Decision maker}&\textbf{Rating}&\textbf{Energy}&\textbf{Quality}&\textbf{Exergy}&\textbf{Entropy}\\
&\textbf{$\tilde{r}$}&\textbf{$\tilde{U}=\tilde{w}.\tilde{r}$}&\textbf{$\tilde{q}$}&\textbf{$\tilde{X}=\tilde{q}.\tilde{U}$}&\textbf{$\tilde{S}=\tilde{U}-\tilde{X}$}\\
\hline
1&(0.10,0.20,0.30)&	(0.07,0.16,0.27)&(0.25,0.40,0.50)&(0.02,0.06,0.14)&(0.05,0.10,0.14)\\
2&(0.20,0.30,0.40)&	(0.14,0.24,0.36)&(0.50,0.60,0.67)&(0.07,0.14,0.24)&(0.07,0.10,0.12)\\
3&(0.30,0.40,0.50)&	(0.21,0.32,0.45)&(0.75,0.80,0.83)&(0.16,0.26,0.38)&(0.05,0.06,0.08)\\
4&(0.70,0.80,0.90)&	(0.49,0.64,0.81)&(0.25,0.40,0.50)&(0.12,0.26,0.41)&(0.37,0.38,0.41)\\
5&(0.70,0.80,0.90)&	(0.49,0.64,0.81)&(0.25,0.40,0.50)&(0.12,0.26,0.41)&(0.37,0.38,0.41)\\ \hline
\textbf{Mean}&\textbf{(0.40,0.50,0.60)}&\textbf{(0.28,0.40,0.54)}&\textbf{(0.40,0.52,0.60)}&\textbf{(0.10,0.20,0.31)}&\textbf{(0.18,0.20,0.23)}\\\hline
\end{tabular}
\end{table}
\end{landscape}
\end{exmp}
Based on the examples studied, we conclude that the use of exergy indicator in place of indicator based on energy will bring more rationality to the decision making process. The use of exergy indicator will enable to account for the quality of the ratings which is neglected in the classical TOPSIS method. 
\section{Evaluation of thermodynamical indicators}
A systematic approach is presented in this section for the ranking of alternatives in MCDM based on exergy indicator in both crisp and fuzzy environment. The MCDM problem consists of $K$ decision makers rating $m$ alternatives on $n$ criteria. The detailed step-by-step procedure is described below:
\subsection{Crisp environment}
\noindent
\textbf{Step 1:} Formulate decision matrices $(D^1,\cdots,D^K)$ for each of the decision maker.
\begin{equation}
D^k = \left [
\begin{array}{cccc}
x_{11}^k&x_{12}^k&\cdots&x_{1n}^k\\
x_{21}^k&x_{22}^k&\cdots&x_{2n}^k\\
\vdots&\vdots&\cdots&\vdots\\
x_{m1}^k&x_{m2}^k&\cdots&x_{mn}^k
\end{array} \right ]
\end{equation}
where $k=1,\cdots,K$ and $x_{ij}^k$ denotes the rating assigned by $k^
{th}$ decision maker to $i^{th}$ alternative for $j^{th}$ criterion .\\
\textbf{Step 2:} Construct the normalized decision matrix $(R^1,\cdots,R^K)$ for each of the decision maker. 
\begin{equation}
R^k=[r_{ij}^k]_{m\text{x}n},\text{ } r_{ij}^k=\left \{ 
\begin{array}{rcl}
\displaystyle \frac{x_{ij}^k}{x_j^{k+}}&&\text{for benefit criterion $j$}\\
\displaystyle \frac{x_{j}^{k-}}{x_{ij}^k}&&\text{for cost criterion $j$}
\end{array} \right.
\end{equation}
where $x_j^{k+}=max_i({x_{ij}^k})$ and $x_j^{k-}=min_i({x_{ij}^k})$ for $i=1,\cdots,m,$ and $j=1,\cdots,n$.\\
\textbf{Step 3:} Construct weight matrix $(W^1,\cdots,W^K)$ for each of the decision maker. 
\begin{equation}
W^k=[w_1^k,\cdots,w_n^k]
\end{equation}
where $w_j^k$ is the weight assigned to $j^{th}$ criterion by $k^{th}$ decision maker.\\
\textbf{Step 4:} Construct energy matrix $(U^1,\cdots,U^K)$ for each of the decision maker. 
\begin{equation}
U^k=\left [
\begin{array}{cccc}
w_1^k.r_{11}^k&w_2^k.r_{12}^k&\cdots&w_n^k.r_{1n}^k\\
w_1^k.r_{21}^k&w_2^k.r_{22}^k&\cdots&w_n^k.r_{2n}^k\\
\vdots&\vdots&\cdots&\vdots\\
w_1^k.r_{m1}^k&w_2^k.r_{m2}^k&\cdots&w_n^k.r_{mn}^k
\end{array} \right ]
\end{equation}
\textbf{Step 5:} Construct quality matrix $(q^1,\cdots,q^K)$ for each of the decision maker.
\begin{equation}
q^k=
\left [
\begin{array}{cccc}
\Big(1-\frac{d(r_{11}^k,\bar{r}_1^k)}{\bar{r}_1^k}\Big)&\Big(1-\frac{d(r_{12}^k,\bar{r}_2^k)}{\bar{r}_2^k}\Big)&\cdots&\Big(1-\frac{d(r_{1n}^k,\bar{r}_n^k)}{\bar{r}_n^k}\Big)\\
\Big(1-\frac{d(r_{21}^k,\bar{r}_1^k)}{\bar{r}_1^k}\Big)&\Big(1-\frac{d(r_{22}^k,\bar{r}_2^k)}{\bar{r}_2^k}\Big)&\cdots&\Big(1-\frac{d(r_{2n}^k,\bar{r}_n^k)}{\bar{r}_n^k}\Big)\\
\vdots&\vdots&\cdots&\vdots\\
\Big(1-\frac{d(r_{m1}^k,\bar{r}_1^k)}{\bar{r}_1^k}\Big)&\Big(1-\frac{d(r_{m2}^k,\bar{r}_2^k)}{\bar{r}_2^k}\Big)&\cdots&\Big(1-\frac{d(r_{mn}^k,\bar{r}_n^k)}{\bar{r}_n^k}\Big)
\end{array} \right ]
\end{equation}
where $\displaystyle \bar{r}_j^k=\frac{1}{m}.(r_{1j}^k+\cdots+r_{mj}^k)$.\\
\textbf{Step 6:} Construct exergy matrix $(X^1,\cdots,X^K)$ for each of the decision maker.
\begin{equation}
X^k=[q_{ij}^k.U_{ij}^k]_{m\text{x}n}
\end{equation}
\textbf{Step 7:} Calculate the average energy and exergy of $i^{th}$ alternative with respect to $k^{th}$ decision maker.
\begin{equation}
\begin{split}
U_i^k&=(U_{i1}^k+U_{i2}^k+\cdots+U_{in}^k)/n\\
X_i^k&=(X_{i1}^k+X_{i2}^k+\cdots+X_{in}^k)/n
\end{split}
\end{equation}
\textbf{Step 8:} Calculate the energy ($U_i$) and exergy ($X_i$) indicators associated with an alternative $i$.
\begin{equation}
\begin{split}
U_i&=(U_{i}^1+U_{i}^2+\cdots+U_{i})^K/K\\
X_i&=(X_{i}^1+X_{i}^2+\cdots+X_{i})^K/K
\end{split}
\end{equation}
\textbf{Step 9:} Calculate the entropy ($S_i$) indicator of an alternative.
\begin{equation}
S_i = U_i-X_i
\end{equation}
\textbf{Step 10:} Rank the alternatives in the order of their exergy indicator. 
\subsection{Fuzzy environment}
The ratings and the weights are assigned in terms of linguistic variables which are then converted to triangular fuzzy numbers. \\
\textbf{Step 1:} Formulate fuzzy decision matrices $(\tilde{D}_1,\cdots,\tilde{D}_K)$ for each of the decision maker.
\begin{equation}
\tilde{D}_k = \left [
\begin{array}{cccc}
\tilde{x}_{11}^k&\tilde{x}_{12}^k&\cdots&\tilde{x}_{1n}^k\\
\tilde{x}_{21}^k&\tilde{x}_{22}^k&\cdots&\tilde{x}_{2n}^k\\
\vdots&\vdots&\cdots&\vdots\\
\tilde{x}_{m1}^k&\tilde{x}_{m2}^k&\cdots&\tilde{x}_{mn}^k
\end{array} \right ]
\end{equation}
where $k=1,\cdots,K$ and $\tilde{x}_{ij}^k=(a_{ij}^k,b_{ij}^k,c_{ij}^k)$ denotes the fuzzy rating assigned by $k^{th}$ decision maker to $i^{th}$ alternative for $j^{th}$ criterion.\\
\textbf{Step 2:} Construct the normalized fuzzy decision matrix $(\tilde{R}^1,\cdots,\tilde{R}^K)$ for each of the decision maker. 
\begin{equation}
\tilde{R}^k=[\tilde{r}_{ij}^k]_{m\text{x}n},\text{ } \tilde{r}_{ij}^k=\left \{ 
\begin{array}{rcl}
\displaystyle (\frac{a_{ij}^k}{c_j^{k+}},\frac{b_{ij}^k}{c_j^{k+}},\frac{c_{ij}^k}{c_j^{k+}})&&\text{for benefit criterion $j$}\\
\displaystyle (\frac{a_{j}^{k-}}{c_{ij}^k},\frac{a_{j}^{k-}}{b_{ij}^k},\frac{a_{j}^{k-}}{a_{ij}^k})&&\text{for cost criterion $j$}
\end{array} \right.
\end{equation}
where $c_j^{k+}=max_i({c_{ij}^k})$ and $a_j^{k-}=min_i({a_{ij}^k})$ for $i=1,\cdots,m,$ and $j=1,\cdots,n$.\\
\textbf{Step 3:} Construct fuzzy weight matrix $(\tilde{W}^1,\cdots,\tilde{W}^K)$ for each of the decision maker. 
\begin{equation}
\tilde{W}^k=[\tilde{w}_1^k,\cdots,\tilde{w}_n^k]
\end{equation}
where $\tilde{w}_j^k=(w_{j1}^k,w_{j2}^k,w_{j3}^k)$ is the weight assigned to $j^{th}$ criterion by $k^{th}$ decision maker.\\
\textbf{Step 4:} Construct fuzzy energy matrix $(\tilde{U}^1,\cdots,\tilde{U}^K)$ for each of the decision maker. 
\begin{equation}
\tilde{U}^k=\left [
\begin{array}{cccc}
\tilde{w}_1^k.\tilde{r}_{11}^k&\tilde{w}_2^k.\tilde{r}_{12}^k&\cdots&\tilde{w}_n^k.\tilde{r}_{1n}^k\\
\tilde{w}_1^k.\tilde{r}_{21}^k&\tilde{w}_2^k.\tilde{r}_{22}^k&\cdots&\tilde{w}_n^k.\tilde{r}_{2n}^k\\
\vdots&\vdots&\cdots&\vdots\\
\tilde{w}_1^k.\tilde{r}_{m1}^k&\tilde{w}_2^k.\tilde{r}_{m2}^k&\cdots&\tilde{w}_n^k.\tilde{r}_{mn}^k
\end{array} \right ]
\end{equation}
\textbf{Step 5:} Construct fuzzy quality matrix $(\tilde{q}^1,\cdots,\tilde{q}^K)$ for each of the decision maker.
\begin{equation}
\tilde{q}^k=
\left [
\begin{array}{cccc}
\Big((1,1,1)-\frac{\tilde{d}(\tilde{r}_{11}^k,\bar{\tilde{r}}_1^k)}{\bar{\tilde{r}}_1^k}\Big)&\Big((1,1,1)-\frac{\tilde{d}(\tilde{r}_{12}^k,\bar{\tilde{r}}_2^k)}{\bar{\tilde{r}}_2^k}\Big)&\cdots&\Big((1,1,1)-\frac{\tilde{d}(\tilde{r}_{1n}^k,\bar{\tilde{r}}_n^k)}{\bar{\tilde{r}}_n^k}\Big)\\
\Big((1,1,1)-\frac{\tilde{d}(\tilde{r}_{21}^k,\bar{\tilde{r}}_1^k)}{\bar{\tilde{r}}_1^k}\Big)&\Big((1,1,1)-\frac{\tilde{d}(\tilde{r}_{22}^k,\bar{\tilde{r}}_2^k)}{\bar{\tilde{r}}_2^k}\Big)&\cdots&\Big((1,1,1)-\frac{\tilde{d}(\tilde{r}_{2n}^k,\bar{\tilde{r}}_n^k)}{\bar{\tilde{r}}_n^k}\Big)\\
\vdots&\vdots&\cdots&\vdots\\
\Big((1,1,1)-\frac{\tilde{d}(\tilde{r}_{m1}^k,\bar{\tilde{r}}_1^k)}{\bar{\tilde{r}}_1^k}\Big)&\Big((1,1,1)-\frac{\tilde{d}(\tilde{r}_{m2}^k,\bar{\tilde{r}}_2^k)}{\bar{\tilde{r}}_2^k}\Big)&\cdots&\Big((1,1,1)-\frac{\tilde{d}(\tilde{r}_{mn}^k,\bar{\tilde{r}}_n^k)}{\bar{\tilde{r}}_n^k}\Big)
\end{array} \right ]
\end{equation}
where $\displaystyle \bar{\tilde{r}}_j^k=\frac{1}{m}.(\tilde{r}_{1j}^k+\cdots+\tilde{r}_{mj}^k)$.\\
\textbf{Step 6:} Construct exergy matrix $(\tilde{X}^1,\cdots,\tilde{X}^K)$ for each of the decision maker.
\begin{equation}
\tilde{X}^k=[\tilde{q}_{ij}^k.\tilde{U}_{ij}^k]_{m\text{x}n}
\end{equation}
\textbf{Step 7:} Calculate the average fuzzy energy and exergy of $i^{th}$ alternative with respect to $k^{th}$ decision maker.
\begin{equation}
\begin{split}
\tilde{U}_i^k&=(\tilde{U}_{i1}^k+\tilde{U}_{i2}^k+\cdots+\tilde{U}_{in}^k)/n\\
\tilde{X}_i^k&=(\tilde{X}_{i1}^k+\tilde{X}_{i2}^k+\cdots+\tilde{X}_{in}^k)/n
\end{split}
\end{equation}
\textbf{Step 8:} Calculate the energy ($U_i$) and exergy ($X_i$) indicator associated with an alternative $i$.
\begin{equation}
\begin{split}
U_i&=(s(\tilde{U}_{i}^1)+s(\tilde{U}_{i}^2)+\cdots+s(\tilde{U}_{i})^K))/K\\
X_i&=(s(\tilde{X}_{i}^1)+s(\tilde{X}_{i}^2)+\cdots+s(\tilde{X}_{i})^K))/K
\end{split}
\end{equation}
where $\displaystyle s(\tilde{x})= \sqrt{\frac{1}{3}(a^2+b^2+c^2)}$.\\
\textbf{Step 9:} Calculate the entropy ($S_i$) indicator of an alternative.
\begin{equation}
S_i = U_i-X_i
\end{equation}
\textbf{Step 10:} Rank the alternatives in the order of their exergy indicator. 

\section{Case studies}
In this section, we take up two examples from the literature to demonstrate the application of the proposed methodology. The ranking based on the thermodynamical indicators is then compared with ranking reported in the literature.
\begin{exmp} \textbf{Human resource selection in crisp environment}\\
This problem is adopted from \cite{shih2007extension}. A company wants to recruit a manager. There are 17 eligible candidates to be evaluated by 4 decision makers (DM) on 7 benefit criteria out of which five are objective and two are subjective. The objective criteria includes language test (C1), professional test (C2), safety rule test (C3), professional skills (C4) and computer skills (C5). The subjective criteria includes panel interview (C6) and one-on-one interview (C7). The score of the candidates in objective and subjective criteria are given in Table~\ref{obj1}. The weights assigned to the different criteria are given in Table~\ref{wt1}.

\begin{table}
\caption{Scores of the candidates for different criteria}
\label{obj1}
\centering
\begin{tabular}{|c|ccccc|cc|cc|cc|cc|}
\hline
\textbf{Candidate}&\multicolumn{5}{|c|}{\textbf{Objective criteria}}&\multicolumn{8}{|c|}{\textbf{Subjective criteria}}\\\cline{2-14}
&&&&&&\multicolumn{2}{|c|}{\textbf{DM1}}&\multicolumn{2}{|c|}{\textbf{DM2}}&\multicolumn{2}{|c|}{\textbf{DM3}}&\multicolumn{2}{|c|}{\textbf{DM4}}\\ \cline{7-14}
&\textbf{C1}&\textbf{C2}&\textbf{C3}&\textbf{C4}&\textbf{C5}&\textbf{C6}&\textbf{C7}&\textbf{C6}&\textbf{C7}&\textbf{C6}&\textbf{C7}&\textbf{C6}&\textbf{C7}\\
\hline
A1&80&70&87&77&76&80&75&85&80&75&70&90&85\\
A2&85&65&76&80&75&\cellcolor{blue!25}65&\cellcolor{blue!25}75&\cellcolor{blue!25}60&\cellcolor{blue!25}70&\cellcolor{blue!25}70&\cellcolor{blue!25}77&\cellcolor{blue!25}60&\cellcolor{blue!25}70\\
A3&78&90&72&80&85&90&85&80&85&80&90&90&95\\
A4&75&84&69&85&65&65&70&55&60&68&72&62&72\\
A5&84&67&60&75&85&\cellcolor{blue!25}75&\cellcolor{blue!25}80&\cellcolor{blue!25}75&\cellcolor{blue!25}80&\cellcolor{blue!25}50&\cellcolor{blue!25}55&\cellcolor{blue!25}70&\cellcolor{blue!25}75\\
A6&85&78&82&81&79&80&80&75&85&77&82&75&75\\
A7&77&83&74&70&71&65&70&70&60&65&72&67&75\\
A8&78&82&72&80&78&70&60&75&65&75&67&82&85\\
A9&85&90&80&88&90&80&85&95&85&90&85&90&92\\
A10&89&75&79&67&77&70&75&75&80&68&78&65&70\\
A11&65&55&68&62&70&50&60&62&65&60&65&65&70\\
A12&70&64&65&65&60&60&65&65&75&50&60&45&50\\
A13&95&80&70&75&70&75&75&80&80&65&75&70&75\\
A14&70&80&79&80&85&80&70&75&72&80&70&75&75\\
A15&60&78&87&70&66&70&65&75&70&65&70&60&65\\
A16&92&85&88&90&85&90&95&92&90&85&80&88&90\\
A17&86&87&80&70&72&80&85&70&75&75&80&70&75\\
\hline
\end{tabular}
\end{table}
\begin{table}
\caption{Weights for different criteria}
\label{wt1}
\centering
\begin{tabular}{cccccccc}
\hline
\textbf{Decision maker}&\textbf{C1}&\textbf{C2}&\textbf{C3}&\textbf{C4}&\textbf{C5}&\textbf{C6}&\textbf{C7}\\ \hline
DM1&0.066&0.196&0.066&0.130&0.130&0.216&0.196\\
DM2&0.042&0.112&0.082&0.176&0.118&0.215&0.255\\
DM3&0.060&0.134&0.051&0.167&0.100&0.203&0.285\\
DM4&0.047&0.109&0.037&0.133&0.081&0.267&0.326\\
\hline
\end{tabular}
\end{table}

\begin{table}
\caption{Thermodynamical indicators and ranking of candidates}
\label{rank1}
\centering
\begin{tabular}{ccccccc}
\hline
\textbf{Candidate}&\textbf{Energy}&\textbf{Exergy}&\textbf{Entropy}&\multicolumn{3}{c}{\textbf{Rank based on}}\\ \cline{5-7}
&\textbf{U}&\textbf{X}&\textbf{S}&\textbf{U}&\textbf{X}&\textbf{Extended TOPSIS}\\ &&&&&&\citep{shih2007extension}\\ \hline
A1&0.860&0.831&0.028&5&\cellcolor{blue!25}6&5\\
A2&0.789&0.771&0.018&\cellcolor{red!25}13&\cellcolor{blue!25}11&14\\
A3&0.934&0.910&0.024&3&3&3\\
A4&0.790&0.768&0.021&12&\cellcolor{blue!25}13&12\\
A5&0.791&0.749&0.042&11&\cellcolor{blue!25}14&11\\
A6&0.873&0.860&0.014&4&4&4\\
A7&0.788&0.770&0.018&\cellcolor{red!25}14&\cellcolor{blue!25}12&13\\
A8&0.836&0.802&0.034&8&\cellcolor{blue!25}9&8\\
A9&0.964&0.946&0.018&2&2&2\\
A10&0.811&0.793&0.018&10&10&10\\
A11&0.690&0.672&0.018&16&16&16\\
A12&0.673&0.632&0.041&17&17&17\\
A13&0.831&0.813&0.017&9&\cellcolor{blue!25}8&9\\
A14&0.849&0.838&0.011&6&\cellcolor{blue!25}5&6\\
A15&0.768&0.748&0.020&15&15&15\\
A16&0.966&0.950&0.017&1&1&1\\
A17&0.846&0.826&0.020&7&7&7\\
\hline
\end{tabular}
\end{table}
The energy, exergy and entropy indicators are evaluated using the procedure described in the previous section. The alternatives are ranked in terms of energy and exergy indicators. The ranking obtained from the thermodynamical indicators is then compared with that reported in \cite{shih2007extension}. The values of the calculated thermodynamical indicators and ranking of the candidates are given in Table \ref{rank1}. The rating which are different from what is reported in the literature are highlighted. It is observed that the energy indicator ranks the alternative almost similar to the ranking based on extended TOPSIS \citep{shih2007extension}. The reason being the terms in the decision matrix of extended TOPSIS are similar to what we defined as energy. The ranking based on exergy indicator is also close to that of extended TOPSIS except for A2 and A5. If we carefully look at the subjective rating of A2 and A5 (highlighted in Table \ref{obj1}), we observe that the variation in the ratings of A5 is more than A2. For A2, the ratings ranges from 60 to 70 for C6 and 70 to 77 for C7. In case of A5, the rating ranges from 50 to 75 for C6 and 55 to 80 for C7. This is also evident from the entropy values of A2 and A5. The quality of the rating reduces with the increase in variations. The confidence in the information that we have depends on its quality. This factor is accounted well if we use exergy indicator.     
\end{exmp}
\begin{exmp} \textbf{Human resource selection in fuzzy environment}\\
This problem is adopted from \cite{chen2000extensions}. A software company wants to hire system analysis engineer. There are three eligible candidates (A1, A2, A3) to be evaluated by three decision makers (DM1, DM2, DM3) on five benefit criteria - emotional steadiness (C1), oral communication skill (C2), personality (C3), past experience (C4) and self-confidence (C5). The ratings and the weights are assigned in terms of linguistic variables. The triangular fuzzy number corresponding to  the linguistic variables for ratings and weights are given in Table \ref{lin_var}. The weights assigned to the different criteria are given in Table \ref{wt2}. The ratings of the three candidates for each of the criteria are given in Table \ref{rat2}. The values of the thermodynamical indicators and ranking of the candidates are given in Table \ref{rank2}.

\begin{table}
\caption{Triangular fuzzy numbers assigned to ratings and weights }
\label{lin_var}
\centering
\begin{tabular}{llcll}
\hline
\multicolumn{2}{c}{\textbf{Ratings}}&&\multicolumn{2}{c}{\textbf{Weights}}\\ \cline{1-2} \cline{4-5}
\textbf{Linguistic variable}&\textbf{Fuzzy number}&&\textbf{Linguistic variable}&\textbf{Fuzzy number}\\ \hline
Very poor (VP)&(0,0,1)&&Very low (VL)&(0,0,0.1)\\
Poor (P)&(0,1,3)&&Low (L)&(0,0.1,0.3)\\
Medium poor (MP)&(1,3,5)&&Medium low (ML)&(0.1,0.3,0.5)\\
Fair (F)&(3,5,7)&&Medium (M)&(0.3,0.5,0.7)\\
Medium good (MG)&(5,7,9)&&Medium high (MH)&(0.5,0.7,0.9)\\
Good (G)&(7,9,10)&&High (H)&(0.7,0.9,1.0)\\
Very good (VG)&(9,10,10)&&Very high (VH)&(0.9,1.0,1.0)\\
\hline
\end{tabular}
\end{table}
\begin{table}
\caption{Weights assigned to different criteria}
\label{wt2}
\centering
\begin{tabular}{lcccc}
\hline
\textbf{Criteria}&\multicolumn{3}{c}{\textbf{Decision maker}}\\ \cline{2-4}
&\textbf{DM1}&\textbf{DM2}&\textbf{DM3}\\ \hline
C1&H&VH&MH\\
C2&VH&VH&VH\\
C3&VH&H&H\\
C4&VH&VH&VH\\
C5&M&MH&MH\\
\hline
\end{tabular}
\end{table}
\begin{table}
\caption{Rating of the candidates for different criteria}
\label{rat2}
\centering
\begin{tabular}{llccc}
\hline
\textbf{Criteria}&\textbf{Candidate}&\multicolumn{3}{c}{\textbf{Decision maker}}\\ \cline{3-5}
&&\textbf{DM1}&\textbf{DM2}&\textbf{DM3}\\ \hline
C1&A1&MG&G&MG\\
&A2&G&G&MG\\
&A3&VG&G&F\\
C2&A1&G&MG&F\\
&A2&VG&VG&VG\\
&A3&MG&G&VG\\
C3&A1&F&G&G\\
&A2&VG&VG&G\\
&A3&G&MG&VG\\
C4&A1&VG&G&VG\\
&A2&VG&VG&VG\\
&A3&G&VG&MG\\
C5&A1&F&F&F\\
&A2&VG&MG&G\\
&A3&G&G&MG\\
\hline
\end{tabular}
\end{table}
\begin{table}
\caption{Thermodynamical indicators and ranking of candidates}
\label{rank2}
\centering
\begin{tabular}{ccccccc}
\hline
\textbf{Candidate}&\textbf{Energy}&\textbf{Exergy}&\textbf{Entropy}&\multicolumn{3}{c}{\textbf{Rank based on}}\\ \cline{5-7}
&\textbf{U}&\textbf{X}&\textbf{S}&\textbf{U}&\textbf{X}&\textbf{Fuzzy TOPSIS}\\
&&&&&&\citep{chen2000extensions}\\ \hline
A1&0.685&0.620&0.065&3&3&3\\
A2&0.825&0.803&0.023&1&1&1\\
A3&0.761&0.683&0.077&2&2&2\\
\hline
\end{tabular}
\end{table}
The ranking of the alternative based on energy and exergy indicators is found to be same as that obtained from fuzzy TOPSIS \citep{chen2000extensions}. The rating of the alternatives are clustered and the variations are less. This proves that when the variations in the ratings are small, same ranking order is obtained from TOPSIS, energy and exergy indicators. An exercise is taken up to highlight the effect of variations on the ranking of candidates based on different methods. The ratings of the candidate A2 by DM1 for criteria C1 and C2 are changed from (G, VG) to (VP, VP), respectively. The new thermodynamical indicators and ranking of the candidates are given in Table \ref{rank2_new}. The variations in the ratings of A2 has resulted in increase in its entropy indicator. The effect of this variation is accounted only in the ranking based on exergy indicator.
\begin{table}
\caption{New thermodynamical indicators and ranking of candidates}
\label{rank2_new}
\centering
\begin{tabular}{ccccccc}
\hline
\textbf{Candidate}&\textbf{Energy}&\textbf{Exergy}&\textbf{Entropy}&\multicolumn{3}{c}{\textbf{Rank based on}}\\ \cline{5-7}
&\textbf{U}&\textbf{X}&\textbf{S}&\textbf{U}&\textbf{X}&\textbf{Fuzzy TOPSIS}\\ \hline
A1&0.685&0.62&0.065&3&\cellcolor{blue!25}2&3\\
A2&0.717&0.591&0.126&2&\cellcolor{blue!25}3&2\\
A3&0.761&0.683&0.077&1&1&1\\
\hline
\end{tabular}
\end{table}
\end{exmp}
The two examples demonstrates the effectiveness of the proposed methodology in both crisp and fuzzy environments. 

\section{Conclusions}
A new model is proposed for MCDM based on thermodynamical analogies. The definition of thermodynamical indicators are derived from the first principles. The energy indicator associated with a rating gives an idea of the quantity of potential energy that a rating carries (based on first law of thermodynamics). The expression of the exergy indicator is derived from the second law of thermodynamics. The exergy indicator gives information on the amount of energy which can be converted to useful work. The entropy indicator gives an idea about the unevenness in the rating of an alternative. The entropy in the present study is defined as the difference between energy and exergy \citep{prathap2011energy} which is different from the Shanon's definition \citep{shannon2015mathematical}. Shanon's entropy assumes a prior distribution for the ratings while no such assumption is made in the present study. The confidence in the ratings reduces if there are large variations in the rating given by different decision makers. The classical TOPSIS method uses what we define as energy for the formulation of the decision matrix. The information on the quality of rating was neglected. We suggest the use of exergy indicator in place of energy to effectively account for the quality of the ratings in the decision making process. The new model is simple to implement and involves less computations compared to TOPSIS. In the proposed model, the alternatives can be ranked directly based on the value of exergy indicator eliminating the calculation of the separation measures from positive and negative ideal solution which is required in TOPSIS.   
\section*{Acknowledgements}
Authors would like to thank and acknowledge the help received from their colleagues of Shock and Vibration Group, CSIR-SERC. This paper is being published with the kind permission of the Director, CSIR-SERC, Chennai.
\section*{References}
\bibliographystyle{elsarticle-harv} 
\bibliography{elsarticle-template-harv}





\end{document}